\documentclass[mlabstract]{jmlr}

\usepackage{tikz}
\newcommand{\circled}[1]{\tikz[baseline=(char.base)]{
    \node[shape=circle,draw=white,fill=black,inner sep=1pt,text=white,font=\scriptsize] (char) {#1};}}

\usepackage{longtable}

\usepackage{booktabs}
\usepackage[load-configurations=version-1]{siunitx} 

\usepackage{lipsum}    
\usepackage{blindtext} 
\usepackage{soul}
\usepackage[blocks]{authblk}

\newcommand{\dddmei}{3D-MEI}
\newcommand{\dddmeis}{3D-MEIs}

\theorembodyfont{\upshape}
\theoremheaderfont{\scshape}
\theorempostheader{:}
\theoremsep{\newline}

\jmlrvolume{}
\firstpageno{1}

\jmlryear{2025}
\jmlrworkshop{Symmetry and Geometry in Neural Representations}

\title{Beyond Pixels: A Differentiable Pipeline for Probing Neuronal Selectivity in 3D} 

\author{%
  \parbox{\textwidth}{\centering
    \normalsize
    \textbf{Pavithra Elumalai}\textsuperscript{1,4},\quad
    \textbf{Mohammad Bashiri}\textsuperscript{5},\quad
    \textbf{Goirik Chakrabarty}\textsuperscript{1,2},\quad
    \textbf{\mbox{Suhas Shrinivasan}}\textsuperscript{1},\quad
    \textbf{Fabian H.~Sinz}\textsuperscript{1-3}\\
    \small
    \textsuperscript{1}Institute for Computer Science, University of Göttingen, Göttingen, Germany\\
    \textsuperscript{2}Campus Institute Data Science (CIDAS), Göttingen, Germany\\
    \textsuperscript{3}Lower Saxony Center for AI \& Causal Methods in Medicine, Göttingen, Germany\\
    \textsuperscript{4} International Max Planck Research School for Intelligent Systems, Tübingen, Germany\\
    \textsuperscript{5}Noselab GmbH, Munich, Germany\\
    \texttt{\{pavithra.elumalai, sinz\}@uni-goettingen.de}
  }%
}
\begin{document}

\maketitle

\begin{abstract}

Visual perception relies on inference of 3D scene properties such as shape, pose, and lighting. To understand how visual sensory neurons enable robust perception, it is crucial to characterize their selectivity to such physically interpretable factors. However, current approaches mainly operate on 2D pixels, making it difficult to isolate selectivity for physical scene properties. To address this limitation, we introduce a differentiable rendering pipeline that optimizes deformable meshes to obtain MEIs directly in 3D. The method parameterizes mesh deformations with radial basis functions and learns offsets and scales that maximize neuronal responses while enforcing geometric regularity. Applied to models of monkey area V4, our approach enables probing neuronal selectivity to interpretable 3D factors such as pose and lighting. This approach bridges inverse graphics with systems neuroscience, offering a way to probe neural selectivity with physically grounded, 3D stimuli beyond conventional pixel-based methods.

\end{abstract}

\section{Introduction}
Primate visual sensory neurons are selective to a wide range of features, from simple edges and luminance to complex attributes such as shape, color, texture, and lighting conditions \citep{hubel1968receptive,hegde2000selectivity, pasupathy2006neural,arcizet2009coding,kim2019neural}. 
Area V4, for example, contains neurons tuned to curvature, material properties, and 3D shape cues, reflecting the growing complexity of representations along the ventral stream \citep{pasupathy2020visual,roe2012toward,pasupathy2019object, srinath2021early}.
Recent work has shown that optimized input images---such as Maximally Exciting Inputs (MEIs)---can reveal aspects of neuronal selectivity and invariances \citep{bashivan2019neural,walker2019inception,Ding2023-ll,Franke2022-do, bashiri2025learning}. 
However, these approaches operate in 2D pixel space, entangling multiple visual factors and making it difficult to isolate interpretable 3D properties. 
To study whether and how neurons show selectivity to physically interpretable factors, we would ideally synthesize the underlying 3D scene factors directly and observe the resulting neuronal responses.

Here, we introduce a differentiable rendering-based pipeline to obtain 3D Maximally Exciting Inputs (\dddmei): an approach that optimizes directly in 3D object space (meshes, textures, poses) through a differentiable renderer. 
This grounds selectivity in physically realizable structure and enables systematic tests of tuning and invariances---such as tolerance to viewpoint, lighting, or material---that pixel-based MEIs do no capture explicitly.
Testing our approach on modeled V4 neurons of the macaque visual cortex, we \circled{1} show that our pipeline can generate physically meaningful objects that resemble the pixel based MEI and strongly drive the neuronal response, \circled{2} demonstrate the potential of our approach to study neuron encoding to complex scene parameters such as lighting conditions.

\section{Methods}

\begin{figure}[t]
  \centering
  \includegraphics[width=1.0\linewidth]{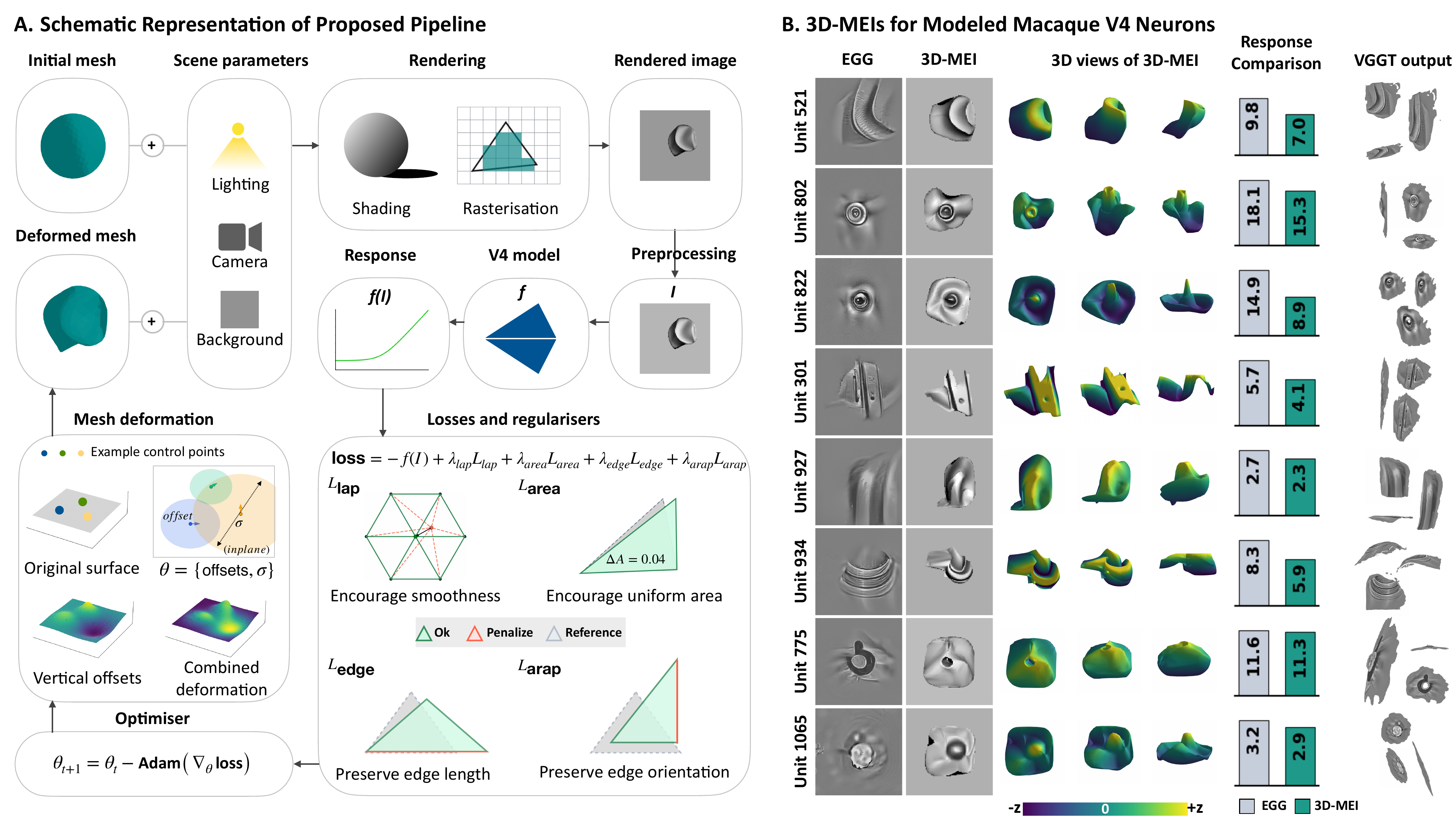}
  \caption{\textbf{A.} Schematic overview of the differentiable rendering pipeline. Starting from an initial mesh and scene parameters (lighting, camera, background), the pipeline iteratively deforms the mesh using radial basis function offsets and optimizes scene parameters to maximize model responses. Losses and regularizers enforce smoothness, uniform surface areas, edge-length preservation, and locally rigid deformations. \textbf{B.} 3D Maximally exciting images (\dddmeis) generated for multiple selected model macaque V4 neurons. For each selected neuron (unit), we show the pixel based EGG MEI and the \dddmei, as well as their 3D views. Right: Bar plots comparing activation values for EGG and \dddmei.  A 2D-to-3D model (VGGT) applied to EGG MEIs fails to generate meaningful 3D structures.}
  \label{fig:Figure 1}
\end{figure}

\label{sec:methods}
Our pipeline (Fig.~\ref{fig:Figure 1}A) operates mainly in 3D object space, deforming an initial mesh to optimize its shape with respect to a neuron's response. 
The neuron's response is obtained using an image-computable response-predictive model.
Given a response-predictive (encoding) model $f$, a differentiable renderer $R$, and a parameterized 3D mesh $M = (V, F)$ with vertices $V$
and faces $F$, 
the pipeline optimizes $V$ such that the rendered projection $I = R(M)$ maximizes the neural model output $f(I)$. 
For $f$, we use deep encoding model from \citet{pierzchlewicz2023energy}, which contains a task driven core with Gaussian readout (see Appendix~\ref{apd:model_description} for details), and use PyTorch3D \citep{ravi2020pytorch3d} as differentiable renderer $R$. 
Meshes are initialized as either a sheet or a sphere, and are placed at the origin while the camera is fixed overhead at a height z at (0, 0, z). 
A fixed lighting is provided by a point light source positioned near the camera (see Fig.~\ref{fig:appA-fig1}D) and the scene is rendered against a uniform gray background. 
$I$ is normalized to match training mean and std, and scaled to have a fixed norm of 25 which was used by \citet{pierzchlewicz2023energy}. The current pipeline does not optimize surface texture and color of objects.

\paragraph{Mesh Deformation via RBF Kernels}
To discourage uncontrolled mesh deformations, we use a set of $K$ radial basis function (RBF) kernels \citep{buhmann2003radial} centered at control points $c_k$ distributed uniformly over the mesh. Each RBF has a learnable scale $\sigma_k$ and offset $\delta_k$, which control the scale and the direction of the displacement, respectively. The deformation $\Delta v$ at any vertex $v$ is then expressed as a weighted sum over all kernel offsets $\Delta v = \sum_{k=1}^{K} \delta_k \cdot \exp\left(-\|v - c_k\|^2/(2\sigma_k^2)\right)$. Here, we use all vertices as control points.

\paragraph{Loss Functions and regularization}
Our goal is to maximize the neuronal response. 
However, unconstrained optimization can yield degenerate meshes. 
To prevent this, we incorporate a set of geometric regularizers. 
Specifically, we use \circled{1} Laplacian smoothing ($\mathcal{L}_{lap}$) \citep{sorkine2004laplacian} to encourage smooth surface curvature, \circled{2} uniform edge length loss ($\mathcal{L}_{edge}$) to penalize large variation in edge lengths of the triangles in the mesh, \circled{3} triangle area loss ($\mathcal{L}_{area}$) to encourage uniform triangle areas to avoid collapse, and finally, \circled{4} As-Rigid-As-Possible Loss ($\mathcal{L}_{arap}$) adapted from \citet{sorkine2007arap} to maintain local surface geometry by penalizing non-rigid deformations. 
Details on the loss function and regularizers can be found in Appendix~\ref{apd:losses}. 
The total loss and optimization objective is
$
\mathcal{L} = -f(I) + \lambda_{lap} \cdot \mathcal{L}_{lap} + \lambda_{edge} \cdot \mathcal{L}_{edge} + \lambda_{area} \cdot \mathcal{L}_{area} + \lambda_{arap} \cdot \mathcal{L}_{arap},
$
which we optimize using the Adam optimizer \citep{kingma2014adam}. $\lambda_{*}$ coefficients were found by a grid search.

\section{Experiments and Results}
\begin{figure}[t]
  \centering
  \includegraphics[width=1.0\linewidth]{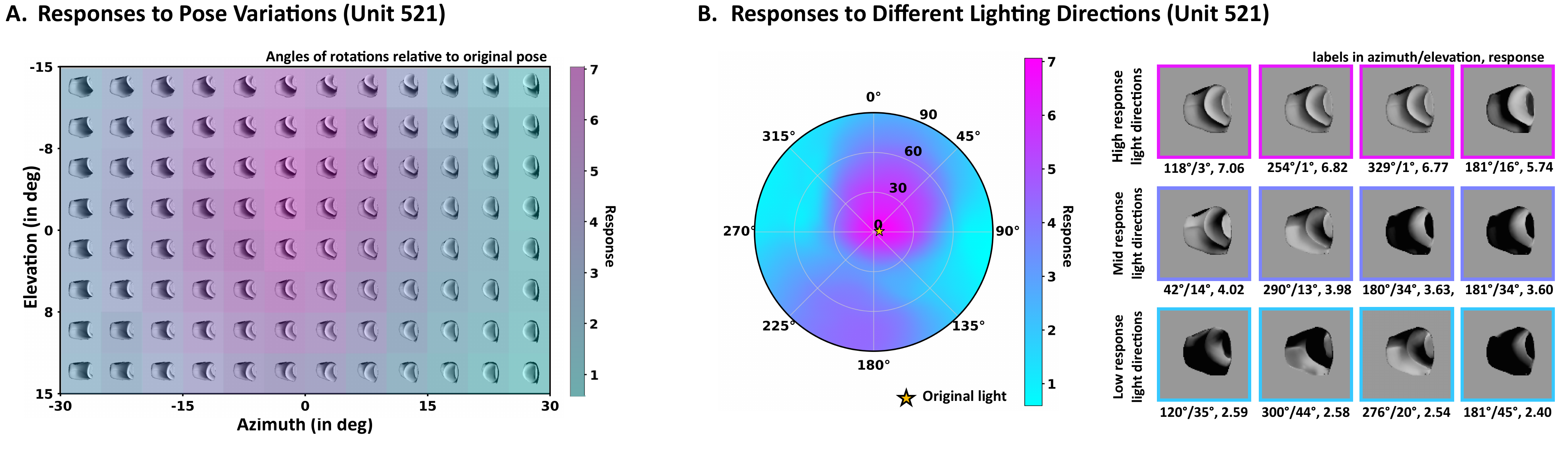}
  \caption{Response profile of Unit 521 to variations in 3D pose and lighting direction.
\textbf{A.} Heatmap of model responses to pose variations (azimuth × elevation) relative to the original view, showing tolerance to moderate rotations.
\textbf{B.} Light-direction tuning of the same unit. Responses are shown for selected light positions on the hemisphere, grouped into high, mid, and low response sets (labels indicate azimuth/elevation in degrees and response values). 
}
  \label{fig:Figure 2}
\end{figure}
First, we verified that our pipeline can recognizably reconstruct known 3D shapes (Fig.~\ref{fig:appA-fig1}A), as well as models of simple and complex cells 
 whose optimal stimuli are pre-defined Gabor patterns (Fig.~\ref{fig:appA-fig1}C). 
 Our pipeline successfully recovers the underlying Gabor structures.
%
For a complex cell, which is phase invariant, the optimization also yields different phase variants of the Gabor under different initialization seeds, which shows the potential of the pipeline to capture invariances.
Next, we apply our pipeline to a set of V4 neurons modeled via deep encoding model  \citep{pierzchlewicz2023energy}. 
We manually selected neurons based on their visual appearance of their pixel optimized MEIs and picked neurons that exhibited geometric, shape-like features. 

The synthesized \dddmeis~closely resemble pixel based MEIs synthesized with energy guided diffusion \citep[EGG,][Fig.~\ref{fig:Figure 1}B]{pierzchlewicz2023energy}. 
Compared to pixel MEIs, \dddmeis~elicit lower activation in the model neurons (Fig.~\ref{fig:Figure 1}B right). 
We attribute this to the fact that the current \dddmeis~do not contain textures, whereas pixel optimized MEIs have texture information. 
This can limit the maximum activation value achievable with the pipeline since V4 neurons are believed to encode both texture and shape \citep{kim2019neural,pasupathy2020visual}. While our approach can be extended to optimize texture and color, we chose to focus on shape optimization for now.
We also compared our method against a 2D-to-3D model such as VGGT \citep{wang2025vggt} that is trained to ``lift'' images of objects to 3D point clouds, by applying it to pixel optimized MEIs. 
However, the resulting geometry lacked meaningful 3D structure (see Fig.~\ref{fig:Figure 1}B).

\paragraph{Exploring Tuning Properties via Scene Manipulation}
Once the mesh is optimized, the scene can be systematically manipulated along physically meaningful axes, enabling interpretable exploration of neuronal tuning properties in 3D space. 
Importantly, these tuning properties cannot easily be explored in images space, since pixel manipulations do not directly capture physical transformations.

To showcase this possibility, we explore tuning of an example neuron along different poses and lighting directions separately, varying one at a time while keeping the other scene parameters fixed.
To assess tuning along pose variations, we systematically rotated the object around azimuth and elevation axes and recorded the model responses (Fig.~\ref{fig:Figure 2}A). 
To assess tuning along lighting directions, we move the point light source along the forward facing half dome and record the model responses for each location of point light (Fig.~\ref{fig:Figure 2}B polar heatmap). For the selected neuron, we observe a strong near-frontal preference, skewed toward the upper-right quadrant of the dome. We also visualize the rendered scenes in three categories: high response, mid response and low response. The curved geometry---a tuning feature in area V4---is visible in almost all light directions but the neuron only prefers certain shadings (Fig.~\ref{fig:Figure 2}B right side) consistent with a shape-light-direction combination that resemble this units preferred stimulus.

\section*{Summary}
We present a proof-of-concept showing that differentiable rendering can probe neural representations in ways inaccessible to pixel-based MEIs.
By operating directly in 3D space, our approach enables the dissection of invariances and sensitivities in relation to physically grounded factors such as shape, curvature, and illumination. 
Our pipeline opens up a new way to probe the inner workings of visual representations, and disentangle tuning of neurons to geometry from other visual features such as texture.
We believe that by optimizing physically plausible 3D stimuli, we can move beyond the limitations of pixel space to understand biological vision.

\acks{
The authors would like to thank the International Max Planck Research School for Intelligent Systems (IMPRS-IS) for supporting Pavithra Elumalai.
Computing time was made available on the high-performance computers HLRN-IV at GWDG at the NHR Center NHR@Göttingen. The center is jointly supported by the Federal Ministry of Education and Research and the state governments participating in the NHR (www.nhr-verein.de/unsere-partner).
This work was funded by the German Federal Ministry of Education and Research (BMBF) via the Collaborative Research in Computational Neuroscience (CRCNS) (FKZ 01GQ2107).
FHS acknowledges the support of the German Research Foundation (DFG): SFB 1233, Robust Vision: Inference Principles and Neural Mechanisms -- Project-ID 276693517, the European Research Council (ERC) under the European Union’s Horizon Europe research and innovation programme (Grant agreement No. 101171526), and  the support of the Lower Saxony Ministry of Science and Culture (MWK) with funds from the Volkswagen Foundation’s zukunft.niedersachsen program (project name: CAIMed - Lower Saxony Center for
Artificial Intelligence and Causal Methods in Medicine; grant number: ZN4257).
}

\bibliography{pmlr-sample}

\newpage
\appendix

\section{Supplementary Results}\label{apd:first}


\begin{figure}[htbp]
  \centering
  \includegraphics[width=1.0\linewidth]{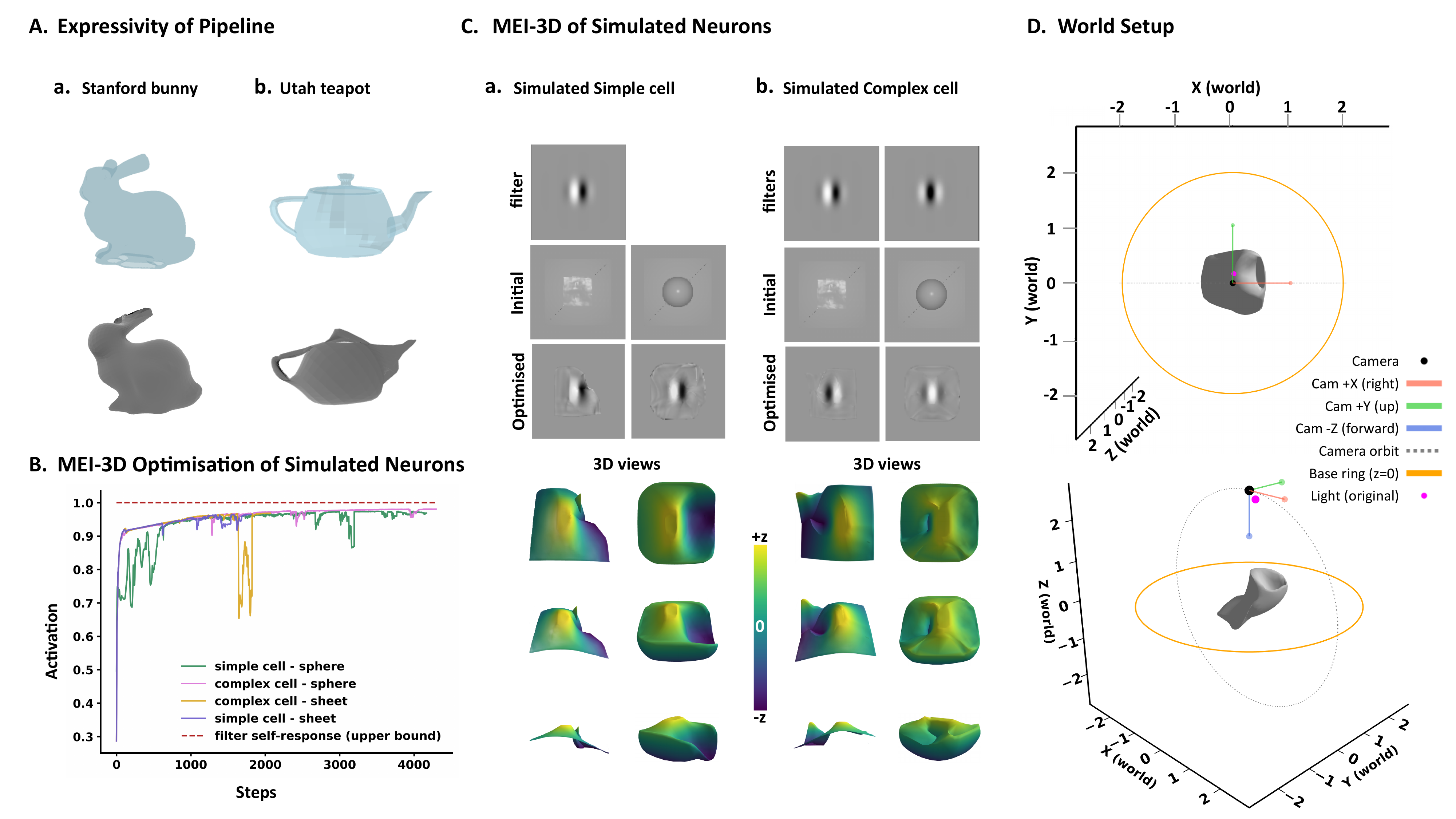}
  \caption{Validation of the pipeline on simulated data.
  \textbf{A.} RBF-based mesh deformation produces complex targets such as the Stanford Bunny (from a sphere) and Utah Teapot (from a torus) by minimizing Chamfer loss \citep{ravi2020pytorch3d} between the input and target meshes.
  \textbf{B.} Optimization curves for simulated simple and complex cells with “sheet” and “sphere” initializations approach to the upper bound of 1.
  \textbf{C.} \dddmei~results for a simulated simple cell \textbf{a.} and complex cell \textbf{b.}. Outputs recover the ground-truth filters; for the phase-invariant complex cell, different initializations yield Gabor-like patterns in distinct phases (example shown).
  \textbf{D.} World setup of the differentiable renderer R, showing world and camera axes. The base ring (z=0) marks the base of half-dome used in light-direction tuning.}
  \label{fig:appA-fig1}
\end{figure}

\section{Details on Loss Function and Regularizers}\label{apd:losses}

Given a response-predictive model $f$, a differentiable renderer $R$, and a parameterized 3D mesh $M = (V, F)$ with vertices $V$,
(where $V = \{v_i\}_{i=1}^{|V|}, \; v_i \in \mathbb{R}^3)$
and faces $F$, 
(where $F = \{f_t=(i,j,k)\}_{t=1}^{|F|}, \; f_t \in \{1,\dots,|V|\}^3)$, 
the pipeline optimizes $V$ such that the rendered projection $I = R(M)$ maximizes the model output $f(I)$. The total loss and optimization objective of our pipeline is
\begin{align*}
\mathcal{L} = -f(I) + \lambda_{lap} \cdot \mathcal{L}_{lap} + \lambda_{edge} \cdot \mathcal{L}_{edge} + \lambda_{area} \cdot \mathcal{L}_{area} + \lambda_{arap} \cdot \mathcal{L}_{arap}
\end{align*}
where, $\mathcal{L}_{lap}$ is the Laplacian smoothing loss, $\mathcal{L}_{edge}$ is uniform edge length loss, $\mathcal{L}_{area}$ is triangle area loss and $\mathcal{L}_{arap}$ is the simplified and adapted version of the classical As Rigid as Possible loss.

We use Laplacian smoothing loss $\mathcal{L}_{lap}$ \citep{sorkine2004laplacian} to enforce local smoothness of the surface by penalizing deviations of each vertex $v_i$ from the mean position of its 1-ring neighbors $\mathcal{N}(i)$.
 \begin{align}
 \mathcal{L}_{\text{lap}} &= \frac{1}{|V|}\sum_{i=1}^{|V|}\left\|\,v_i - \frac{1}{|\mathcal N(i)|}\sum_{j\in\mathcal N(i)} v_j \right\|^{2}
 \end{align}

We define uniform edge length loss $\mathcal{L}_\text{edge}$ to discourage edges from becoming disproportionately long or short compared to others. Without it, the mesh could stretch, shear, or collapse. By penalizing variance in edge lengths, the mesh maintains spatial regularity and stable deformations.

\begin{align}
\mathcal{L}_{\text{edge}} &= \frac{1}{|E|}\sum_{(i,j)\in E}\Big(\left\|v_i-v_j\right\| - \bar \ell\Big)^2,
\quad \bar \ell = \frac{1}{|E|}\sum_{(i,j)\in E}\left\|v_i-v_j\right\|
\end{align}
where, $E$ is the edge set extracted from $F$.

Similarly, triangle area loss $\mathcal{L}_{\text{area}}$ enforces uniformity in the areas of all the triangular faces. As the vertices move, some triangles can become skinny or even degenerate. By penalizing the variance in the area of triangles, $\mathcal{L}_{\text{area}}$ promotes evenly sized triangles, preserving mesh quality and preventing local collapse.

\begin{align}
\mathcal{L}_{\text{area}} &= \frac{1}{|F|}\sum_{t\in F}\big(A_t - \bar A\big)^2,
\quad A_t = \tfrac{1}{2}\left\|(v_j-v_i)\times(v_k-v_i)\right\|, \quad
\bar A = \frac{1}{|F|}\sum_{t\in F} A_t
\end{align}
%


While uniform edge length loss and triangle area losses regularize local distances and face sizes, they do not directly constrain changes in edge orientation. To address this, we use a \textit{direction-preserving} variant of the As-Rigid-As-Possible (ARAP) energy \citep{sorkine2007arap} that penalizes deviations of each deformed edge from its original direction. Specifically, each edge vector $v_i - v_j$ is projected onto its initial unit vector $\widehat e_{ij}^{(0)} = \widehat{\,v_i^{(0)} - v_j^{(0)}\,}$, and we penalize the residual orthogonal component:
\begin{equation}
\mathcal{L}_{\text{arap}} = \frac{1}{|E|}\sum_{(i,j)\in E}\left\|(v_i - v_j) - \text{proj}_{\hat e^{(0)}_{ij}}(v_i - v_j)\right\|^2 
\end{equation}
This preserves the original orientation of the edges while ignoring changes in magnitude.



\section{Response predictive V4 model}\label{apd:model_description}
We use the ``Gaussian model'' defined in \citet{pierzchlewicz2023energy} as the image computable encoding model for macaque V4 neurons. The model contains a pre-trained robust ResNet50 ($L_{2}, \varepsilon = 0.1$) \citep{he2016deep,salman2020adversarially} core and a Gaussian readout \citep{lurz2020generalization}. The model uses 3 layers with 1024 channels, resulting in a 1024 dimension feature space, followed by batch normalization\citep{ioffe2015batch} and ReLU non-linearity. 
The EGG based pixel optimized MEIs shown in Fig \ref{fig:Figure 1} is also generated from this model.
\end{document}